\documentclass[lettersize,journal]{IEEEtran}
\usepackage{amsmath,amsfonts}
\usepackage{algorithmic}
\usepackage{algorithm}
\usepackage{array}
\usepackage[caption=false,font=normalsize,labelfont=sf,textfont=sf]{subfig}
\usepackage{textcomp}
\usepackage{stfloats}
\usepackage{url}
\usepackage{verbatim}
\usepackage{graphicx}
\usepackage{cite}
\usepackage{bm}
\usepackage{colortbl}
\usepackage{makecell}

\usepackage[colorlinks = True,
linkcolor=black,            
anchorcolor=black,
urlcolor=black,            
citecolor=black]{hyperref}

\usepackage{multicol}
\usepackage{multirow}
\usepackage{marvosym}
\usepackage{booktabs}
\usepackage{pifont}
\usepackage{color,xcolor}
\definecolor{light-gray}{gray}{0.929}

\begin{document}

\title{ScaleFlow++: Robust and Accurate Estimation of 3D Motion from Video}

	\author{Han~Ling,
	Yinghui~Sun,
	Quansen~Sun
	and
	Yuhui Zheng,~\IEEEmembership{Member,~IEEE,}
	\IEEEcompsocitemizethanks{\IEEEcompsocthanksitem Han Ling and Quansen Sun are with the School of Nanjing University of Science and Technology, Nanjing, Jiangsu; Yinghui Sun is with the School of Southeast University; Yuhui Zheng is with the Nanjing University of Information Science and Technology. (Corresponding authors: Quansen Sun) \protect\\
		E-mail: 321106010190@njust.edu.cn, sunyh@seu.edu.cn,
		sunquansen@njust.edu.cn, zhengyh@vip.126.com}  
\thanks{This paper was produced by the IEEE Publication Technology Group. They are in Piscataway, NJ.}
\thanks{Manuscript received April 19, 2021; revised August 16, 2021.}}

\markboth{Journal of \LaTeX\ Class Files,~Vol.~14, No.~8, August~2021}%
{Shell \MakeLowercase{\textit{et al.}}: A Sample Article Using IEEEtran.cls for IEEE Journals}


\maketitle

\begin{abstract}
Perceiving and understanding 3D motion is a core technology in fields such as autonomous driving, robots, and motion prediction. This paper proposes a 3D motion perception method called ScaleFlow++ that is easy to generalize. With just a pair of RGB images, ScaleFlow++ can robustly estimate optical flow and motion-in-depth (MID).
Most existing methods directly regress MID from two RGB frames or optical flow, resulting in inaccurate and unstable results. Our key insight is cross-scale matching, which extracts deep motion clues by matching objects in pairs of images at different scales. Unlike previous methods, ScaleFlow++ integrates optical flow and MID estimation into a unified architecture, estimating optical flow and MID end-to-end based on feature matching. Moreover, we also proposed modules such as global initialization network, global iterative optimizer, and hybrid training pipeline to integrate global motion information, reduce the number of iterations, and prevent overfitting during training.
On KITTI, ScaleFlow++ achieved the best monocular scene flow estimation performance, reducing SF-all from 6.21 to 5.79. The evaluation of MID even surpasses RGBD-based methods. In addition, ScaleFlow++ has achieved stunning zero-shot generalization performance in both rigid and nonrigid scenes. Code is available at \url{https://github.com/HanLingsgjk/CSCV}. 
\end{abstract}


\section{Introduction}
\IEEEPARstart{T}{hree-dimensional} motion estimation aims to estimate the motion trend of pixels in a pair of images in 3D space. It has essential applications in video understanding\cite{kong2022human,byrne_expansion_2009}, autonomous driving\cite{noauthor_future_nodate,mori_first_2013,9351818,brebion2021real,sormoli2024optical,shi2023panoflow,alletto2018self}, and autonomous robot navigation\cite{marinho_guaranteed_2018}.
For example, in autonomous driving scenarios\cite{9284628,liu2022camliflow}, the car platform needs to obtain the 3D motion state of typical targets to make path decisions. Most existing platforms use Lidar to obtain 3D point clouds of scenes and construct corresponding relationships between two frames of 3D point clouds to estimate 3D motion. However, due to limited range and sparse results, this active depth perception scheme cannot effectively handle objects at a distance and abnormally reflective surfaces. In addition, the expensive price and maintenance costs further limit the broader application of Lidar. In this paper, we consider using a monocular camera to estimate 3D motion in dynamic scenes (commonly called normalized scene flow), which is a more stable and easy-to-maintain sensor that is not limited by limited range and sparse results.

\textbf{Challenge:} Normalized scene flow (NSF) consists of optical flow and motion-in-depth. Monocular cameras can robustly estimate 2D optical flow and normalized camera motion trajectories by constructing pixel-by-pixel feature associations between two frames\cite{jiang2021learning,zhang20013d}. However, monocular cameras cannot provide real scene depth\cite{luo2020consistent}. In dynamic scenes, 2D optical flow is a function of scene speed, camera motion trajectory, and scene depth. Given only the optical flow and camera trajectory, solving these three parts is an under-constrained problem.

\begin{figure}[!t]
	\centering
	\includegraphics[width=3.2in]{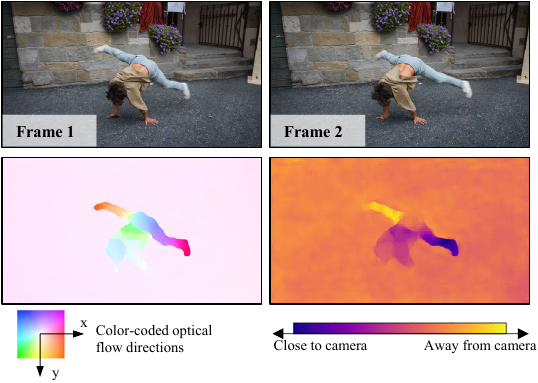}
	
	\caption{\textbf{Zero-shot Generalization in Real Scenes.} Only needing to input two consecutive frames of images, ScaleFlow++ can robustly estimate dense 3D motion fields. The lower left corner of the image is the color-coded optical flow map, and the lower right corner is the motion-in-depth (MID). MID provides additional depth cues in the Z-axis direction, such as the dancer's left leg moving away from the camera and the right leg moving closer to the camera.}
	\label{fig_dance}
\end{figure}
\textbf{Previous work:} In early time-to-collision (TTC) work\cite{camus1995calculating,muller2009time,sagrebin2008robust}, people found that there is a close correlation between the scale change of the object and the depth motion. They estimated the scale change of a specific image block through algorithms such as SIFT\cite{lowe_distinctive_2004}, and converted the scale change into sparse depth motion through the transformer formula. Recent works\cite{yang_upgrading_2020,badki_binary_2021} regress dense scale changes from optical flow fields and RGB images through deep networks. However, their over-reliance on regression leads to lower accuracy and unstable results. Moreover, they separate the calculation of optical flow and motion-in-depth, requiring the network to be trained in stages rather than end-to-end training.

\begin{figure*}[!t]
	\centering
	\includegraphics[width=6.6in]{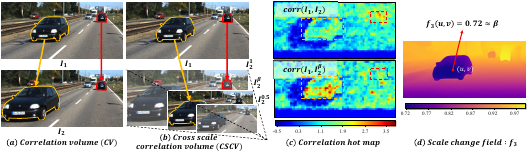}
	
	\caption{Cross-scale matching idea. We match cars between two consecutive frames $I_1$ and $I_2$, where the yellow car is close to the camera and the red car is relatively static to the camera. (a) The CV module in the optical flow baseline\cite{teed_raft_2020} matches cars at the same scale, while the yellow car cannot match well because of the scale change. (b) CSCV matches objects in the 3D scale space, so that each object can achieve the perfect matching of position and scale simultaneously. 
		(c) We visualize the correlation hot map sampled from the CV module. $corr(I_1,I_2^\beta)$ means that CV is built based on $I_1$ and $I_2$, where the meaning of each pixel point $(x, y)$ is the correlation between the poin $I_1(x, y) $ and its corresponding point $I_2^\beta (\beta(x+\bm{f}(x,y)), \beta(y+\bm{f}(x,y)))$, $\bm{f}$ is the ground truth optical flow. \textbf{The correlation of the scale-changed yellow car in $corr(I_1,I_2)$ is smaller than that of the scale-invariant red car, while the correlation of the yellow car in $corr(I_1,I_2^\beta)$ is higher, which proves that the scale change has an important impact on the stability of optical flow matching}. 
		(d) Dense scale change field $f_3$ estimated from our CSCV, the value of each pixel represents the scale change ratio.}
	\label{fig_sim}
\end{figure*}

In addition to the regression dependency dilemma faced by motion-in-depth estimation, existing optical flow pipelines are also plagued by scale issue. Most optical flow methods\cite{teed_raft_2020,sun2018pwc,brox2004high,papenberg2006highly,brox2010large} rely on feature-matching techniques to associate objects in two frames by comparing the correlations between features. However, previous studies have pointed out that the commonly used CNN feature extractors only satisfy translation invariance and do not satisfy scale invariance\cite{zeiler2014visualizing}. When an object moves along the Z-axis (depth direction) of the camera, its scale change often leads to feature-matching failure. We demonstrate the significant impact of scale variation on feature matching in Fig.~\ref{fig_sim}(c).

\textbf{Our method:} 
Fortunately, previous work has already explored many issues related to scale. As early as 1998, Lin et al.\cite{lindeberg1998feature}. proposed a scheme based on the Laplace operator to calculate the maximum response position in the scale pyramid to solve the problem of scale change in feature recognition. Even if the scale of the input image changes, the output response features are stable. Furthermore, in 2005, Mikolajczyk\cite{mikolajczyk2005comparison} successfully estimated the scale changes of the bell tower captured at different focal lengths using a similar approach.
Therefore, inspired by the success of previous work, we propose a cross-scale correlation volume (CSCV) module. As shown in Fig.~\ref{fig_sim}, CSCV extends the original optical flow matching from the 2D-pixel plane to the scale pyramid, matching the correct features at the correct position and scale, improving the previous dilemma of estimating MID overly dependent on regression.

We propose an enhanced monocular 3D flow estimation network ScaleFlow++ based on the successful optical flow framework RAFT\cite{teed_raft_2020}. Its core feature is to replace the original 4D correlation volume with CSCV, and estimate both optical flow and MID based on feature matching in a unified architecture. To enhance the information aggregation capability of the network, we propose a lightweight global iterative optimizer (GIR). In previous studies, most RAFT-style methods used the GRU optimizer. However, this simple single-scale optimizer can only provide a local view, which makes the optimizer overly focused on the correlation features obtained from local sampling and unable to coordinate global motion information over a larger scale range. On the contrary, GIR referred to Unet's\cite{huang2020unet} successful experience and constructed an hourglass-shaped multi-scale residual convolutional network. In the encoding stage, it used the ConvNextV2\cite{woo2023convnext} module with a super large convolution kernel and employed pyramid pooling\cite{he2015spatial} at the minimum scale to coordinate global information. The results indicate that using the GIR optimizer can significantly improve background performance and slightly enhance foreground performance while keeping the total time almost constant.

In addition to GIR, we also introduced a global initialization module, which is mainly due to the task alienation observed in RAFT-style methods at different iteration periods. In the early stages of iteration, the optimizer starts optimizing from zero, which requires the optimizer to make aggressive estimates. At the end of the iteration, the optimizer focuses more on fine-tuning. The difficulty and focus of tasks vary significantly at different stages of iteration. Therefore, in order to alleviate the training burden of task alienation on the optimizer, we used an additional optimizer to quickly initialize the entire motion field at a scale of 1/16,  to alleviate the alienation problem in subsequent iterations and reduce the total number of iterations.

During the training phase, we improved the commonly used truth supervision mode and proposed an enhanced training pipeline that combines self-supervised mixed truth. Specifically, we generate a flying foreground with random textures and shapes, iteratively overlay it on the original image pairs, and calculate the 3D stream truth of the flying foreground as a self-supervised training label. Previous works\cite{teed_raft_2020} often used solid color blocks to cover the source image in order to actively create occlusion. However, this behavior encourages the network to overly rely on regressing to estimate optical flow, resulting in poor generalization ability. Our hybrid training pipeline actively creates occlusions while forcing the network to learn 3D flow of randomly moving foreground through texture matching, ultimately balancing occlusion prediction and texture matching learning.

The significant performance improvement validates our motivation. In the KITTI\cite{Menze2015ISA} scene flow test set, ScaleFlow++ surpassed all NSF methods, reducing the core metric SF-all from 6.21\% to 5.79\%. In MID estimation, the minimum MID error was reduced from 42.84 to 38.44. In the Sintel\cite{butler2012naturalistic} optical flow test set, ScaleFlow++ outperforms most similar methods with highly competitive performance. Moreover, as shown in Fig.~\ref{fig_dance}, ScaleFlow++ also has strong generalization ability and robustness. In Sec.~\ref{sec66}, we demonstrated the outstanding performance of ScaleFlow++ in unseen scenarios.

\textbf{Differences from the preliminary version:}
This work is a substantial extension of our previous paper, Scale-flow\cite{ling2022scale}, presented at the ACM MM 2022 conference. The preliminary work proposes a monocular 3D flow estimation method based on cross-scale matching, which alleviates the previous dilemma of relying solely on regression to calculate MID. This work comprehensively refactored the previous method while retaining the core module CSCV and made substantial improvements in multiple aspects, such as sports field initialization, iterative optimizer, and training methods, reducing the core indicator MID error from 48.9 to 38.44. The improved version significantly improved performance on various test benchmarks. The new contributions can be summarized as follows:
(1) We introduce a self-supervised and truth mixed training pipeline that helps the network better balance the learning between occlusion fitting and texture matching, alleviate potential overfitting problems during training, and better predict motion prospects.
(2) In order to solve the problem of standard GRU iteration modules having difficulty perceiving global motion, we propose an iterative optimizer GIR with a global perspective, which significantly improves the background performance of the network and slightly enhances the foreground performance while keeping the total time unchanged.
(3) To alleviate the task alienation problem of the optimizer at different iteration stages, we constructed an initialization module to initialize the 3D flow field quickly. Reducing the total number of iterations while improving the optimizer's optimization capability.
(4) We further demonstrated the superiority of our method in the field of optical flow on the Sintel dataset.
(5) Based on the self-supervised generalization method ADF, we provide a real-world generalization model for readers to test.
Our code and test model can be obtained on \url{https://github.com/HanLingsgjk/CSCV}.
\section{Related Works}
In this section, we first introduce the optical flow task and scene flow task closely related to 3D flow, and then present the motion-in-depth estimation and cross-scale matching technique, which is the primary source of inspiration for the CSCV module.

\subsection{Optical Flow}
Early methods\cite{sun_learning_2008,horn_determining_1981} viewed optical flow as an optimization problem for minimizing energy. Based on the regularization term and the constant brightness assumption, the optical flow results are optimized iteratively. Some recent excellent works have inherited their ideas\cite{teed_raft_2020,jiang_learning_2021,teed_raft-3d_2021}, gradually updating the optical flow field through correlation matching and warping techniques. Compared with the previous methods \cite{yang_volumetric_2019,sun2018pwc,ranjan_optical_2017,lu_devon_2020,ilg_flownet_2017,dosovitskiy_flownet_2015} of directly using convolutional networks to regress optical flow, the method based on correlation matching has a clearer theoretical background and more robust results.
In this work, inspired by the successful experience of correlation matching, we extended this matching pattern to the scale dimension based on CSCV, improving the scale robustness of the optical flow baseline model.

It is worth mentioning that although some previous optical flow works\cite{xu_scale_2012,wang_autoscaler_2016,teed_raft_2020} also used the idea of scale, our method differs fundamentally from theirs. Previous works are often based on multi-scale technology to increase the network receptive field, improve the inference speed, and strengthen the loss function. However, it is still image-to-image matching in essence. CSCV introduces a pattern for optical flow matching in scale space, a brand-new attempt.

\subsection{Scene Flow}
Scene flow aims to estimate the motion of pixels in 3D space, which is the main component of the 3D flow problem.

\textbf{Methods Based on Rigid Assumption.}
Early methods\cite{menze_object_2015,vogel_3d_2015} decompose the image into rigid blocks, and iteratively optimize the 3D motion of the rigid blocks based on the rigid assumption and regularization terms.
In recent scene flow methods\cite{teed_raft-3d_2021,liu2022camliflow,yang_learning_2021} based on convolutional networks, a semantic network is often introduced to judge whether pixels belong to the same rigid object, and the 3D flow is optimized again according to the semantic results. Although it performs well on datasets, such methods are difficult to promote and apply in field scenarios due to the need for depth information and additional semantic labels.
In addition, sub-modules such as 3D flow and semantic estimation are independent. Any defects in sub-module will affect overall performance, as the entire pipeline depends on its results.

\textbf{Methods Based on Point Cloud.}
Point cloud based methods\cite{puy2020flot,liu2019flownet3d,wei2021pv,wu2019deep} capture 3D point clouds of the surrounding environment through an active laser or stereo camera, and establish the corresponding relationship between two frame point clouds to estimate 3D motion. However, this kind of method is limited by the sparse results and limited range, which is challenging to apply in more common video scenes.
There is also a point cloud method based on the fusion framework, Camliflow\cite{liu2022camliflow}. By highly fusing the point cloud and the optical flow pipeline, the two pipelines can obtain complementary results at the same time. Camliflow achieves end-to-end scene flow estimation, but it still has the problem of depth information dependence.

\begin{figure*}[h]
	\centering
	\includegraphics[width=1.0\textwidth]{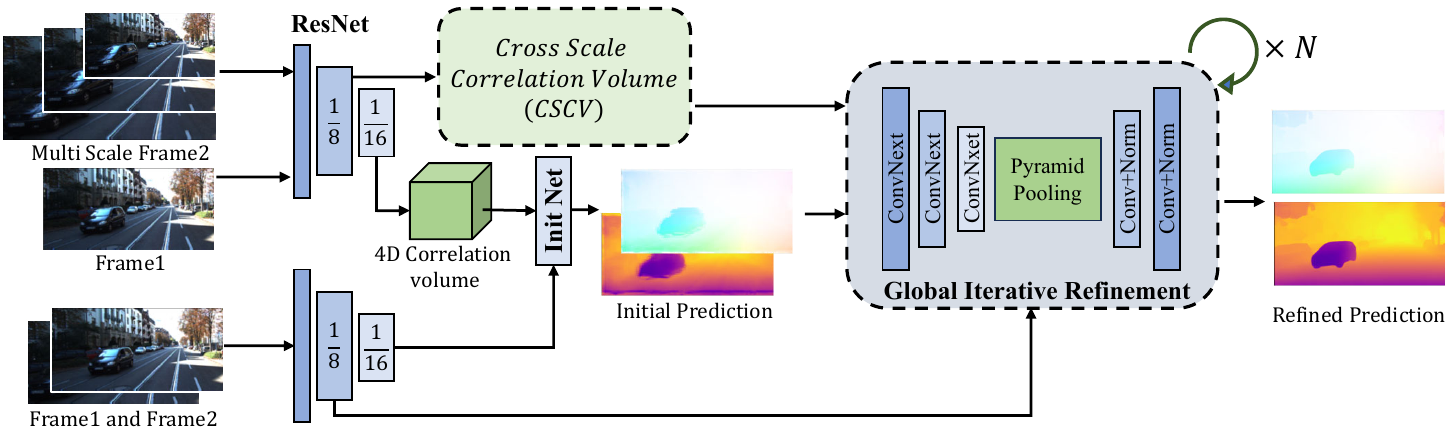}
	\caption{Overview of ScaleFlow++. The network is mainly divided into two stages: initialization and iterative optimization. In the initialization stage, we construct a 4D correlation volume based on the 1/16 features extracted from ResNet and sample it using an all-zero optical flow field. Based on this, we regress the initialized 3D motion field. In the iterative optimization phase, we sample cross-scale correlation features from CSCV based on the current motion field, encode them, and send them to the GIR module for optimization. Output the final optimization result after N iterations.}
	\label{methodfig}
\end{figure*}
\subsection{Motion-in-depth (MID)}
MID describes the motion in the depth direction through the scale change of depth. It has many practical application scenarios, such as time-to-collision (TTC) estimation, LiDAR scene flow, scene flow, Etc. MID is also the core indicator of our ScaleFlow++.

Early TTC works\cite{camus1995calculating,muller2009time,sagrebin2008robust} estimated the MID by tracking the motion trajectory of the interest points and building a motion model. Because the distribution of the interest points is random, this kind of method often makes it hard to pay attention to the primary target, yielding only sparse and low-precision results.
Binary TTC\cite{badki_binary_2021}  is a successful TTC scheme proposed recently, which can regress a dense MID field segmentally.
Binary TTC sends the scaled image to the encoder to directly return the MID result of a certain range, and the specific range depends on the scaling factor.
Therefore, a complete MID estimation requires multiple regressions to cover the complete MID interval. Unlike them, ScaleFlow++ only needs one calculation to get complete and dense MID results.

Yang \& Ramanan\cite{yang_upgrading_2020} proposed an other monocular normalized scene flow method. They obtained dense MID results based on optical flow step-wise regression optical expansion and MID. However, the same as previous TTC work, they separated optical flow and MID estimation, resulting in the depth motion estimation result dependent on the optical flow result. Unlike them, ScaleFlow++ integrates optical flow and depth motion estimation into a harmonious task based on CSCV, making information blend and help each other. In addition, the cross-scale matching mechanism can also help the optical flow module more accurately estimate the objects with scale changes.
\subsection{Cross-scale Matching}
Cross-scale matching aims to establish the correspondence between objects in different frames, and the scales of objects in these frames often have large changes, which is a difficult problem faced by current optical flow methods.
In the early works, SIFT\cite{lowe_distinctive_2004} obtains scale-invariant features in the scale space through hand-designed feature extraction operators, and then matches objects of different scales based on these stable features.
In addition to being used for matching, cross-scale matching technology can also be used to estimate the scale changes of objects between frames. As early as 2005, Mikolajczyk\cite{mikolajczyk2005comparison} used Lindeberg's scale selection method\cite{lindeberg1998feature} to successfully estimate the optical expansion of clock towers captured at different focal lengths. Inspired by the above works, CSCV performs optical flow matching in scale space and estimates a dense scale change field based on the cross-scale matching results.

\begin{figure*}[h]
	\centering
	\includegraphics[width=1.0\textwidth]{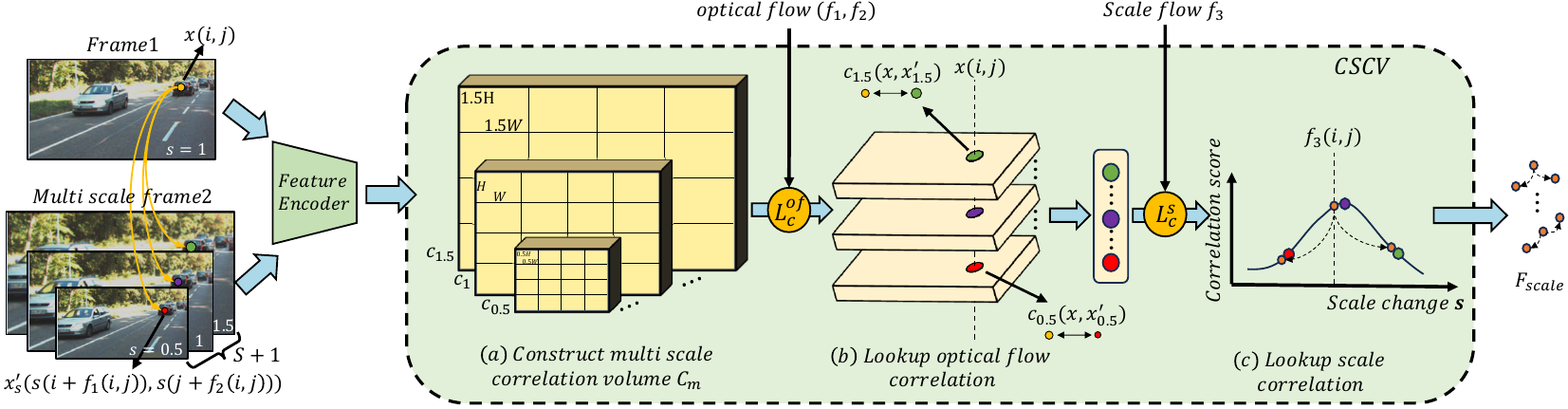}
	\caption{The complete structure of cross-scale correlation volume (CSCV). The CSCV module is composed of three parts: multi scale correlation volume $C_m$, optical flow lookup operator $L_c^{of}$, and scale lookup operator $L_c^{s}$. After the optical flow field $(f_1,f_2)$ and scale change field $f_3$ are input, the optical flow feature $F_{of}$ and multi-scale optical flow feature $F_{multi}$ are first sampled from $C_m$ by the $L_c^{of}$ operator, while the scale feature $F_{scale}$ is obtained from $F_{multi}$ by the $L_c^{s}$ operator.}
	\label{cscv}
\end{figure*}

\section{Method}\label{method}

The framework of ScaleFlow++ is shown in Fig.~\ref{methodfig}, similar to RAFT, which constructs a Cross-Scale Correlation Volume (CSCV) on all pixel pairs and iteratively optimizes the 3D motion field through an GIR optimizer. We also constructed a small 4D correlation volume based on features of 1/16 size for rapid initialization of optical flow and scale flow, thereby reducing the optimization difficulty of the GIR. In the inference process, GIR \textbf{directly outputs the optical flow $(f_1,f_2)$ and scale change $f_3$ fields for CSCV sampling}. Combining previous work\cite{yang_upgrading_2020}, we convert $f_3$ into motion-in-depth, which is convenient for training and can be more widely applied to downstream tasks.

Below, we will provide a detailed introduction to the construction and training process of ScaleFlow++, as well as how to convert network estimation results into a 3D motion field.

\subsection{Feature Extraction}
We replaced the residual network in RAFT with ResNet18\cite{He_2016_CVPR}, which has the advantage of being able to use pre-trained weights from MAE\cite{he2022masked} or ImageNet\cite{krizhevsky2012imagenet}. The results indicate that this can improve training efficiency and introduce additional prior knowledge to improve generalization performance. Specifically, we constructed a feature encoder and a context encoder, respectively:

\textbf{Feature Encoder.} 
For the feature encoder $g_{\theta}$, we take its truncated outputs at 1/8 and 1/16 scales, where $g_{\theta}^1:\mathbb{R}^{H\times W\times 3}\rightarrow \mathbb{R}^{(H/8)\times (W/8)\times D}$ and $g_{\theta}^2:\mathbb{R}^{H\times W\times 3}\rightarrow \mathbb{R}^{(H/16)\times (W/16)\times D} (D=256)$.

\textbf{Context Encoder.}
Following RAFT, we build a context encoder $h_{\theta}$, whose input is the superposition of frames $I_1$ and $I_2$, and the rest of the structure is the same as $g_{\theta}$. The output of the 1/16 scale $F_{init}(D=384)$ is directly used to initialize, while the output of the 1/8 scale is divided into two parts: one is the context feature $F_{context}(D=192)$ and the other is the implicit feature $h_0 (D=192)$ used for GIR iteration.

\subsection{Initialization of Optical Flow and Scale Flow}\label{sec32}
RAFT-type methods usually initialize the flow field to 0 and then refine it iteratively through a GRU module. However, we found that this leads to two problems. Firstly, GRU iterative networks must simultaneously complete tasks such as large field optimization (early iteration) and fine-tuning (late iteration), significantly increasing the difficulty of training the iterative network. Secondly, due to limitations in memory and cost, it is not easy to increase the sampling range of the correlation volume during iteration in order to obtain a larger receptive field.

In ScaleFlow++, we drew on the successful experiences of IGEV\cite{xu2023iterative} and SEA-RAFT\cite{wang2024sea} and used an independent lightweight network to quickly initialize the 3D flow, reducing the number of iterations of the optimizer and alleviating task differences at different iteration stages. Specifically, the input for initializing the network is the contextual feature $F_{init}$ and the correlation feature $F_{of}^{init}$. We first use the GIR module proposed in Sec.~\ref{gir} to encode the above features:
\begin{equation}
	h_{init}=\textbf{\textrm{GIR}}(F_{of}^{init},F_{init})
\end{equation}
where $F_{of}^{init}\in \mathbb{R}^{H\times W\times(2r+1)\times (2r+1)}$ is sampled from a 1/16 scale 4D correlation volume, the sampling optical flow field is all 0, and the sampling radius $r$ is 6.

Next, we upsample the hi to 1/8 scale and use a simple two-layer convolution head to predict the initialized optical flow field $(f_1^0,f_2^0)$ and scale change field $f_3^0$:
\begin{equation}
	\begin{split}
		(f_1^0,f_2^0) &=\textrm{Conv}(\textrm{ReLU}(\textrm{Conv}(\textrm{Upsample}(h_{init}))))\\
		f_3^0 &=\textrm{Conv}(\textrm{ReLU}(\textrm{Conv}(\textrm{Upsample}(h_{init}))))
		\label{eq:eq6}
	\end{split}
\end{equation}
\subsection{Cross-scale Correlation Volume (CSCV)}

In this part, we introduce the construction of the cross-scale correlation volume (CSCV) module. As shown in Fig.\ref{cscv} , firstly, the correlation volume $C_m$ is constructed from the dot multiplication of input multi-scale features. Then the multi-scale optical flow correlation features $F_{multi}$ and scale correlation features $F_{scale}$ are extracted from $C_m$ successively using the optical flow lookup operator $L_c^{of}$ and scale lookup operator $L_c^{s}$. 

\subsubsection{The Input of CSCV}
The main function of CSCV is to sample the correlation features of the current input 3D flow field. Specifically, the input 3D flow field is composed of an optical flow field $(f_1,f_2)$ and a scale change field $f_3$. The optical flow field describes the displacement of pixels between two frames, and the scale change field is the reciprocal of optical expansion (OE), which describes the pixel scaling factor of the second frame for perfect matching the first frame.

When constructing $C_m$, multi-scale image features are extracted by a residual convolutional network $g_{\theta}$, where the input of the network is $I_1$ and $P_{I_2}$:
\begin{equation}
	P_{I_2}= \{I_2^s \mid s =  0.5, 0.5+1/S,...,1,..., 1.5\}
\end{equation}
where $I_2^{s}\in \mathbb{R}^{(8\times H\times s)\times (8\times W\times s)\times 3}$, $S$ specifies the degree of refinement of the input scale, which is generally set to 4 in our experiment. Because subsequent iterations were conducted on the 1/8 scale, the width and height were multiplied by a coefficient of 8 to facilitate the expression of following functions.

\begin{figure*}[h]
	\centering
	\includegraphics[width=6in]{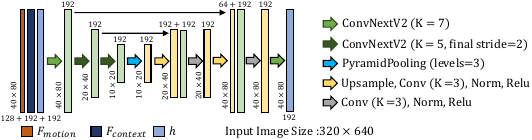}
	\caption{Global Iterative Refinement Module (GIR). Similar to Unet, GIR is mainly composed of funnel-shaped encoders and decoders. Specifically, GIR uses ConNextV2 with large kernel convolution characteristics at multiple scales, where a single convolution kernel can cover almost the entire frame. Combined with the pyramid pooling module at the smallest scale, it greatly enhances global perception capability.}
	\label{fig_GIR}
\end{figure*}

\begin{figure}
	\centering
	\includegraphics[width=3.4in]{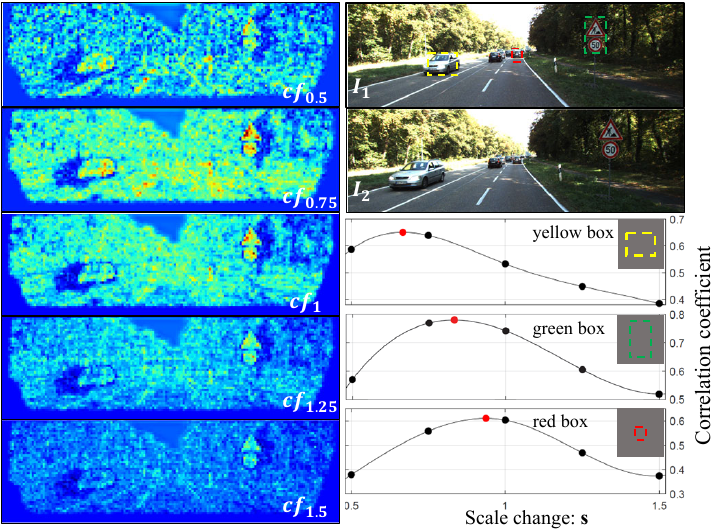}
	\caption{Visualization of the scale matching by CSCV. On the left is the multi-scale optical flow feature $F_{multi}$ when $S=4$ and $r_f=0$. The redder the color of the point, the higher the correlation. On the right is the correlation curve, where the value of each black point is the mean value of the points in the box on the left, and the red point is the extreme value obtained by  interpolation. The extreme points of the cars in the yellow box and the signs in the green box tend to be 0.5 and 0.75, which means they are moving toward the camera rapidly. The extreme point of the red box tends to 1, which means that the object is relatively stationary with the camera.}
	\label{fig_sm}
\end{figure}

\subsubsection{Multi Scale Correlation Volume}
The multi scale correlation volume $C_m$ is a collection of multiple 4D correlation matrices. The construction of the 4D matrix follows RAFT, which can be sampled by the coordinates of a pair of matching points. The sample result is the correlation of these two points.
Specifically, for the given image features $g_{\theta}^1(I_2^s)\in \mathbb{R}^{(H \times s)\times (W \times s)\times D}$ and $g_{\theta}^1(I_1)\in \mathbb{R}^{H\times W\times D}$ , dot them to obtain a single 4D correlation matrix $c_s$:
\begin{equation}
	\begin{gathered}
		c_{s}\left(g_{\theta}^1\left(I_{2}^{s}\right), g_{\theta}^1\left(I_{1}\right)\right) \in \mathbb{R}^{H \times W \times(H \times s) \times(W \times s)} \\
		(c_s)_{i j k l}=\sum_{h} g_{\theta}^1\left(I_{2}^{s}\right)_{i j h} \cdot g_{\theta}^1\left(I_{1}\right)_{k l h}
		\label{eq:eq5}
	\end{gathered}
\end{equation}
where $(i,j)$, $(k,l)$ are the coordinates of two pairs of points on $I_2^s$ and $I_1$ respectively, $h$ is the dimension index of the feature, and $c_s$ is the 4D correlation volume between $I_1$ and $I_2$ with scale $s$.

The final $C_m$ is a set about $c_s$:
\begin{equation}
	C_m= \{c_s \mid s =  0.5, 0.5+1/S,...,1,..., 1.5\}
\end{equation}

\subsubsection{Optical Flow Lookup Operator}
We define the optical flow lookup operator $L_c^{of}$ to sample the correlation features corresponding to the current optical flow field $(f_1,f_2)$ from $C_m$. 

Given an optical flow field $(f_1,f_2)$, it maps each pixel $x=(k,l)$ in $I_1$ to $x'=s(k+f_1(k,l),l+f_2(k,l))=(i,j)$ in $I_2^s$, $L_c^{of}$ samples in the neighborhood $N_{x'}^f$ of $x'$:
\begin{equation}
	N_{x'}^f = \{\ {(i+d_u,j+d_v) \mid d_u,d_v\in \{\ -r_f,...,r_f\}\ } \}\
\end{equation}
where $r_f$ is an integer, it specifies the size of the sampled neighborhood, and neighborhood sampling provides the optimal direction for the network.

Based on the sampling coordinates provided by $N_{x'}^f$, we conducted bilinear interpolation sampling for each $c_s$ in $C_m$ to get the collection of multi-scale optical flow correlation feature $F_{of}$:
\begin{equation}
	F_{of} =\{ cf_s \mid s =  0.5, 0.5+1/S,...,1,..., 1.5  \} 
\end{equation}
where $cf_s \in \mathbb{R}^{H\times W\times(2r_f+1)\times (2r_f+1)}$. In Fig.~\ref{fig_sm}, we demonstrate the working principle of $F_{of}$ through a simplified cross-scale matching example.

Finally, we spliced all the elements in $F_{of}$ to form a 5D correlation matrix $F_{multi} \in \mathbb{R}^{H\times W\times(2r_f+1)\times (2r_f+1) \times (S+1)}$, The newly added fifth dimension is the scale dimension, which can be used to sample the correlation between the $I_2$ features of different scales and the $I_1$ features of the original scale.

\subsubsection{Scale Lookup Operator.}
We define the scale lookup operator $L_c^{s}$ to sample the correlation features corresponding to the current scale change field $f_3$ from $F_{multi}$.

Given a scale change field $f_3$, it describes the scale change of a pixel block between two frames. The scale change field $f_3$ maps each pixel $x=(k,l)$ in $I_1$ to $x'=f_3(k,l) \times(k+f_1(k,l),l+f_2(k,l))=(i,j)$ in $I_2^{f_3(k,l)}$, $L_c^{s}$ sample the neighborhood $N_{x'}^s$ in the scale direction of point $x'$:
\begin{equation}
	N_{x'}^s = \{\ {f_3(k,l)+d_s \mid d_s\in \{\ -r_s,...,r_s\}\ } \}\
\end{equation}
where $r_s$ is a real number, we set $r_s=1/S$ in our experiments, $N_{x'}^s$ provides optimal awareness in the scale direction.

Based on the sampling coordinates provided by $N_{x'}^s$, we sample in $F_{multi}$ based on bilinear interpolation, and splice the results into scale correlation feature $F_{scale} \in \mathbb{R}^{H\times W\times(2r_f+1)\times (2r_f+1)\times (2r_s+1)} $.
Moreover, we also follow the RAFT strategy to extract the optical flow correlation feature $F_{of}\in \mathbb{R}^{H\times W\times(2r_f+1)\times (2r_f+1)\times 4}$, which is sampled from $c_1$ ( $2 \times2$ pooling for three times) by $L_c^{of}$.

\subsection{Global Iterative Refinement} \label{gir}

\subsubsection{Optimizer}
In most RAFT-type methods, the optimizer's field of view is often limited to the size of the search radius in the correlation volume, and can not obtain a global field of view. To overcome this problem, we propose the Global Iterative Refinement Module (GIR), which has a detailed structure as shown in Fig.\ref{fig_GIR}. Similar to Unet, GIR consists of funnel-shaped encoders and decoders, and the encoding stage mainly uses ConvNextV2 with a large field of view. Unlike the commonly used GRU iteration module, multi-scale large kernel convolution and pyramid pooling module can ensure that GIR obtains sufficient global perception capability. 

\subsubsection{Initial State}
The initial states of the optical flow field $(f_1^0,f_2^0)$ and scale change field $f_3^0$ are given by the initialization network in Sec.~\ref{sec32}, and the initial hidden layer features $h_0$ is given by the context encoder.

\subsubsection{Updating}
We iteratively optimize the 3D flow field based on residual update $f^{k+1} = f^k + \Delta f$. Before starting the update, we first need to encode the correlation features $F_{scale}, F_{of}$, as well as the optical flow field $(f_1,f_2)$ and scale change field $f_3$, into motion features $F_{motion}$:
\begin{equation}
	F_{motion}=\textbf{\textrm{MotionEncoder}}(F_{scale},F_{of},f_1,f_2,f_3)
\end{equation}
where \textbf{\textrm{MotionEncoder}} is an encoder composed of a simple two-layer convolution, $F_{motion}\in \mathbb{R}^{H\times W\times 128}$.

Next, as shown in Fig.~\ref{fig_GIR}, use the GIR module to iteratively update the hidden features:
\begin{equation}
	h_t=\textbf{\textrm{GIR}}(F_{motion},F_{context},h_{t-1})
\end{equation}

Finally, the residual $\Delta f=(\Delta f_1,\Delta f_2,\Delta f_3)$ for updating is estimated from $h_t$ by a simple two-level convolution head:

\begin{equation}
	\begin{split}
		(\Delta f_1,\Delta f_2) &=\textrm{Conv}(\textrm{ReLU}(\textrm{Conv}(h_t)))\\
		\Delta f_3 &=\tanh(\textrm{Conv}(\textrm{ReLU}(\textrm{Conv}(h_t))))
		\label{eq:eq11}
	\end{split}
\end{equation}

Considering that the scale flow requires more detailed estimation, we performed an additional update on $f_3$ individually using the GIR network at the end.

\begin{figure*}[!t]
	\centering
	\includegraphics[width=7.1in]{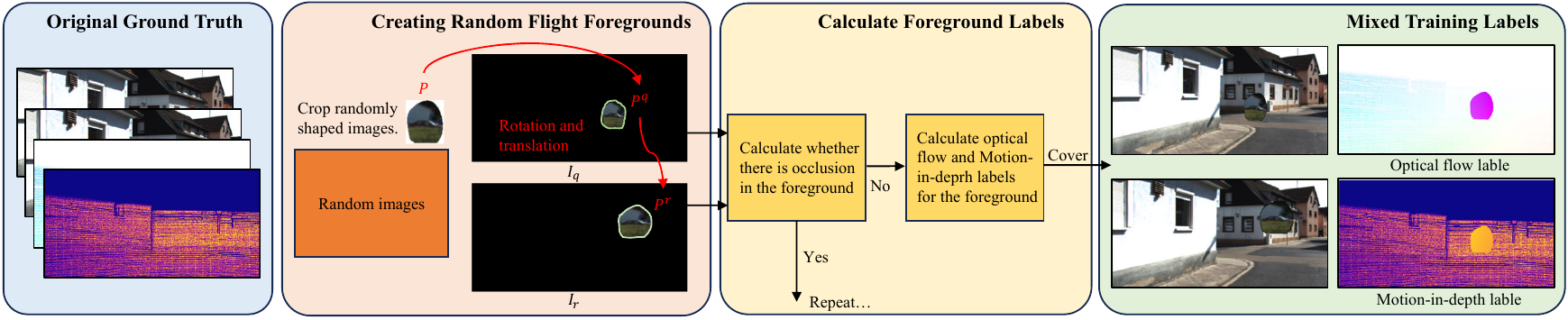}
	
	\caption{Hybrid training data generation pipeline. Our pipeline is divided into three parts. Firstly, we crop randomly shaped foreground image blocks from random images and perform spatial rotation and translation transformations between two frames. Afterwards, it is determined whether the image block has been occluded (flying out of the image range). If no occlusion has occurred, the corresponding optical flow truth value and depth change truth value are calculated. Finally, iteratively add the generated flight foreground to the original image and update the related truth labels at the same time.}
	\label{fig_mix}
\end{figure*}

\subsection{Hybrid Training Pipeline} \label{6d}
We use a combination of self-supervised and ground truth methods to train our model. In this section, we focus on introducing the self-supervised data generation pipeline. As shown in Fig. \ref{fig_mix}, details are as follows.

\textbf{Creating Random Flight Foregrounds.}
Firstly, based on the randomly generated closed curve (generated by the Bessel curve), crop the corresponding image block $p$ on a random image, and then project it to the $P$ in the camera coordinate system:
\begin{equation}
	p(i) =(x,y,1) 
\end{equation}
\begin{equation}
	P(i) = Z\bm{K^{-1}}p(i)= (X,Y,Z) 
\end{equation}
where $i$ is the point contained in $p$, $\bm{K}$ is the camera intrinsic matrix, and the depth $Z$ of all pixels in $P$ is 1.

After obtaining $P$, we first perform the initial spatial transformation to obtain $P^q$, and then project it onto the pixel coordinate system to get the $p^q$ on image $I_q$:
\begin{equation}
	P^q(i) =\bm{R}_r P(i)+\bm{T}_r
\end{equation}
\begin{equation}
	p_q(i) =\bm{K}P^q(i)
\end{equation}
among them, $\bm{R}_r$ and $\bm{T}_r$ are the rotation matrix and translation matrix of random values. Similarly, we perform the spatial transformation on $P^q$ to obtain $P^r$ and project it onto the pixel plane to get $p^r$ on image $I_r$.

Note that the captured image blocks here are random textures, as we found that using foreground flying objects with clear semantic information (such as complete car, human, and animal contours) does not perform as well as random textures.
Specifically, when occlusion occurs, the network becomes overly confident in inferring the motion of the occluded part of the object as planar motion. Random textures can encourage networks to calculate the motion of these flying foreground based on texture matching, and balance the learning weight with occlusion regression.

\textbf{Calculate Foreground Labels.}
Before calculating the label for the flight foreground, it is necessary first to determine whether there is an occlusion in the foreground. Because the flight foreground is primarily a random texture, the network cannot make reasonable inferences about the occluded parts. We propose a difference rate $Df$ indicator to measure whether the foreground is retained:
\begin{equation}
	Df = \frac{|N_r-N_q|}{|N_r+N_q|}
\end{equation}
where $N_q$ and $N_r$ represent the number of foreground flying object pixels in the first and second frames, respectively. In the experiment, we only used flight prospects with $Df$ less than 0.5 to ensure no significant obstructions or great scale changes during the flight.

After filtering, calculate the accurate value of the flight foreground optical flow and the true value of the depth change rate as follows:
\begin{equation}
	\bm{f_{fg}} = p_r - p_q
\end{equation}
\begin{equation}
	\bm{\tau_{fg}} = Z^r / Z^q
\end{equation}
where $\bm{f_{fg}}$ is the optical flow of the flight foreground, and $\bm{\tau_{fg}}$ is the corresponding motion-in-depth.

\textbf{Optical Flow Loss.}
We supervise the optical flow results with $L_1$ distance between prediction and ground truth, the optical flow loss $\mathcal{L}_f$ is defined as:
\begin{equation}
	\mathcal{L}_f=\sum_{k=1}^{N} \gamma^{N-k} (\left\|f_1^k-f_{1 gt}\right\|_{1}+\left\|f_2^k-f_{2 gt}\right\|_{1})
\end{equation}
we set $\gamma=0.8$, $N=6$ in our experiments, the $\bm{f_{gt}}=(f_{1gt},f_{2gt})$ is composed of a hybrid of truth labels from the dataset and the generated $\bm{f_{fg}}$ labels.

\textbf{Scale Change Field Loss.}
According to Eq (\ref{eq:eq1}), the Scale change field $f_3$ equal to MID $\bm{\tau}$ under the slight rotation assumption, which allows us to use the scene flow datasets to train the $f_3$. The loss is defined as follows:
\begin{equation}
	\mathcal{L}_s=\sum_{k=1}^{M} \gamma^{M-k}\left\|f_3^{k}-\bm{\tau}_{gt}\right\|_{1}
\end{equation}
where $\bm{\tau}_{gt}$ is also a hybrid of true values $\bm{\tau}$ and generated labels $\bm{\tau}_{fg}$,  $\bm{\tau} = Z'_{gt} / Z_{gt}$, $Z'_{gt}$ and $Z_{gt}$ are the ground truth depths of matching pixels
in the second frame and the first frame, we set $\gamma=0.8$, $N=7$ in our experiments.

The overall loss function is:
\begin{equation}
	\mathcal{L}=\mathcal{L}_f + \mathcal{L}_s
\end{equation}

\subsection{Convert ScaleFlow++ output to 3D motion } \label{oemid}
Referring to previous work, we converted the 3D motion $(f_1,f_2,f_3)$ into motion-in-depth and scene flow.
Firstly, based on yang's derivation\cite{yang_upgrading_2020}, we can obtain the following approximate equation:
\begin{equation}
	f_3\approx\frac{Z'}{Z}=\bm{\tau}
	\label{eq:eq1}
\end{equation}
where, $Z$ and $Z'$ is the depth of the object in two frames, and $\bm{\tau}$ is the motion-in-depth.

Furthermore, based on Eq.\ref{eq:eq1}, we can derive the relationship between $(f_1,f_2,f_3)$ and scene flow $\bm{t}$:
\begin{equation}
	\bm{t}=Z\bm{K^{-1}}[(f_3-1)\bm{Cd}+f_3(f_1,f_2)]=Z\overline{\bm{t}}
	\label{eq:eq3}
\end{equation}
where  $\bm{Cd}$ is the image coordinate index matrix,  normalize scene flow $\overline{\bm{t}}=\bm{K^{-1}}[(f_3(\bm{p})-1)\bm{p}+f_3(\bm{p})\bm{f}]$. We have presented a detailed derivation process in the \textbf{Supplementary Material}.

\begin{table*}[htbp]
	\centering
	\caption{\textbf{Training Details.} \textit{Pre} represents the pre-training stage, and \textit{Ft} represents the fine-tuning stage. The hybrid training pipeline is used by default during training, and the number of flying foreground additions is 1. Hybrid represents the number of foreground in training}
    \begin{tabular}{ccccccccc}
    \toprule
	Section & Dataset & Stage & Batch & Refine Iteration & Train Iteration & Size  & LR    & Hybrid \\
	\midrule
	Ablation (\ref{ex:ab}) & D+ S-18 &       & 4     & 6     & 10k   & 320x720 & 0.000125 & 1 \\
	\midrule
	\multirow{2}[2]{*}{Motion-in-depth (\ref{ex:mid})} & D+T+S+M+K-160 & \textit{Pre}   & 4     & 6     & 30k   & 320x720 & 0.00025 & 1 \\
	& K-160 & \textit{Ft}    & 6     & 6     & 5k    & 320x896 & 0.00005 & 2 \\
	\midrule
	\multirow{2}[1]{*}{Sintel-test (\ref{ex:op})} & D+T+S+M+K15 & \textit{Pre}   & 4     & 6     & 30k   & 320x720 & 0.00025 & 1 \\
	& T+S+M & \textit{Ft}    & 6     & 6     & 30k   & 320x896 & 0.00005 & 2 \\
	\midrule
	\multirow{2}[1]{*}{Kitti-test (\ref{ex:scene})} & D+T+S+M+K15 & \textit{Pre}   & 4     & 6     & 30k   & 320x720 & 0.00025 & 1 \\
	& K15   & \textit{Ft}    & 6     & 6     & 5k    & 320x896 & 0.00005 & 2 \\
	\midrule
	Generalization (\ref{sec66}) & ADFactory\cite{ling2024adfactory}   &       & 4     & 6     & 30k   & 320x768 & 0.00025 & 1 \\
	\bottomrule
	\end{tabular}%
	\label{tab:trainset}%
\end{table*}%

\section{Experiments}
In this section, we introduce the key details of the dataset and experiment, and then further discuss the experimental results. Specifically, the experiment includes the following aspects. Firstly, a comprehensive ablation was conducted on the proposed module. Secondly, test the performance of ScaleFlow++ on mainstream monocular 3D task MID. Thirdly, submit the results of ScaleFlow++ to public testing platform and compared with the most advanced scene flow and optical flow methods currently available. Finally, we qualitatively evaluate the generalization performance of ScaleFlow++ in unseen scenarios.

\subsection{Datasets and Training Setting}
We mainly compared the methods quantitatively on the KITTI\cite{Menze2015ISA} and Sintel\cite{butler2012naturalistic} baselines, as they have publicly available online evaluation websites and have been widely applied in previous works. 

\textbf{Datasets Used:} 200 image pairs from KITTI 2015 (K15)\cite{Menze2015ISA}. Commonly used pre-training datasets Flyingthings3D (T)\cite{mayer2016large}, Driving (D)\cite{mayer2016large}, Monkey (M)\cite{mayer2016large}  and Sintel (S)\cite{butler2012naturalistic} from movie.

\textbf{Dataset Split:} For the convenience of ablation experiments, we also divided the Sintel dataset into two subsets. One is the training set (S-18) containing 824 image pairs from 18 scenes, and the other is the testing set (S-5) containing 240 image pairs from 5 scenes. In addition, for the MID task, we use the same strategy as yang\cite{yang_upgrading_2020} to split the dataset, selecting one out of every five images in KITTI for evaluation (K-40) and the remaining 160 images for fine-tuning training (K-160).

\textbf{Training Setting:} Tab.~\ref{tab:trainset} shows the specific training process, during which we also performed enhancement operations such as scaling, translation, and color domain transformation on the image pairs. Please refer to RAFT\cite{teed_raft_2020} for details.

\subsection{Ablation}\label{ex:ab}
In this section, we discuss the effectiveness of the various components of the ScaleFlow++ method proposed in this paper. The complete ablation results are shown in Tab.~\ref{tab:ablation}; we gradually stack the proposed modules based on the Base model.
\begin{table*}[h]
	\centering
	\caption{\textbf{Ablation Experiment.} Resnet\textsuperscript{*} represents the use of ImageNet weights for initialization. $Mid$ is the motion-in-depth error, and the calculation method is detailed in Eq.~\ref{eq:eqmid}. EPE is the endpoint error of optical flow, and ALL is the outlier rate. }
	\begin{tabular}{cllccccccc}
		\toprule
		&       &       & \multicolumn{2}{c}{S-5-clean} & \multicolumn{2}{c}{S-5-final} & \multicolumn{3}{c}{K15} \\
		\midrule
		& Ablation & Iter & EPE $\downarrow$   & ALL $\downarrow$   & EPE $\downarrow$   & ALL $\downarrow$    & EPE $\downarrow$    & ALL $\downarrow$    & $Mid \downarrow$ \\
		\midrule
		A     & Base  & 10k   & 2.70  & 11.43  & 3.23  & 15.01  & 4.89  & 18.95  & 167.86  \\
		
		B     & Base+Ht & 10k   & 2.46  & 10.78  & 3.17  & 14.88  & 4.63  & 17.57  & 153.98  \\
		
		C     & Base+Ht+CSCV & 10k   & 2.35  & 10.03  & 2.98  & 13.53  & 4.28  & 16.12  & 143.66  \\
		
		D     & Base+Ht+CSCV+Resnet & 10k   & 2.35  & 10.00  & 3.17  & 14.07  & 4.38  & 16.47  & 137.56  \\
		
		E     & Base+Ht+CSCV+Resnet\textsuperscript{*}  & 10k   & 2.11  & 8.42  & 2.81  & 11.69  & 4.96  & 17.13  & 175.84  \\
		F     & Base+Ht+CSCV+Resnet\textsuperscript{*}+RNN  & 10k   & 2.06  & 8.11  & 2.63  & 11.76  & 3.51  & 11.85  & 127.55  \\
		
		G     & Base+Ht+CSCV+Resnet\textsuperscript{*}+GIR & 10k   & 1.87  & 7.59  & 2.53  & 10.51  & 3.60  & 11.91  & 115.03  \\
		\midrule
		\multirow{2}[0]{*}{H} & \multirow{2}[0]{*}{Base+Ht+CSCV+Resnet\textsuperscript{*}+GIR+Init} & 10k   & 1.82  & 7.09  & 2.29  & 9.77  & 3.17  & 10.31  & 114.78  \\
		&       & 15k   & 1.72  & 6.66  & 2.23  & 9.24  & 3.23  & 10.20  & 112.73  \\
		\midrule
		\multirow{2}[1]{*}{I} & \multirow{2}[1]{*}{Base+Ht+CSCV+Resnet+GIR+Init} & 10k   & 2.04  & 8.32  & 2.77  & 11.66  & 3.15  & 10.83  & 95.52  \\
		&       & 15k   & 1.89  & 7.68  & 2.57  & 10.12  & 3.26  & 10.61  & 96.07  \\
		\bottomrule
	\end{tabular}%
	\label{tab:ablation}%
\end{table*}

\textbf{Base Model:}
We used the previous version Scale-flow as the ablation baseline and replaced the CSCV with the 4D Correlation Volume in RAFT. The rest remains unchanged, using the residual feature extractor and GRU iterative optimizer from RAFT.

\textbf{Hybrid Training Pipeline (Ht):} As shown in rows A and B of Tab.~\ref{tab:ablation}, the mixed training pipeline Ht comprehensively improves the performance of all indicators. This proves that our proposed hybrid training pipeline can effectively enhance the scene flow performance of the network without increasing training costs. 

\textbf{Cross-Scale Matching (CSCV):}
As shown in rows B and C of Table~\ref{tab:ablation}, the model's optical flow and scale flow performance have been significantly improved. Especially in the driving scenario KITTI with significant changes in scale, the error of optical flow decreased from 4.63 to 4.28, which proves the effectiveness of our cross-scale matching strategy.

\textbf{ResNet and Weight Pre-training:} 
In rows D and E of Table~\ref{tab:ablation}, we replaced the original residual convolution with standard Resnet and used pre-trained weights from ImageNet. The experimental results show that the performance of Resnet is close to the accuracy of the RAFT original feature extractor, but after using pre-trained weights, the performance of the Sintel dataset is significantly improved, while the KITTI dataset is significantly reduced. This is easy to understand because there are almost no driving scenes in ImageNet, and its data domain is closer to Sintel. We also verified the relationship between training iteration and pre-training weights. As shown in H and I, the method using pre-training weights still maintained significant advantages when more iterations were being trained.

The above experiment proves that appropriate pre-training weights can effectively improve model performance in similar data domains, and this advantage will continue to be maintained with increasing iteration times. Therefore, we recommend using ImageNet weights only on the Sintel dataset in subsequent comparisons.

\begin{figure}[!t]
	\centering
	\includegraphics[width=3.2in]{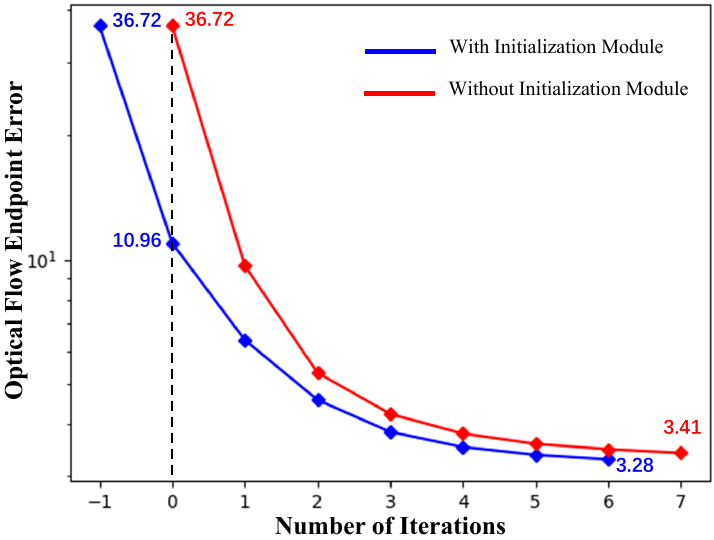}
	
	\caption{Initialization Module. We selected the first 100 sequences of K15 to calculate the average optical flow endpoint error (EPE) for each iteration. The blue and red colors in the figure correspond to H and G in Table~\ref{tab:ablation}, respectively. The difference is that the red has an additional GIR iteration to ensure fair comparison. When the optical flow field is all 0, the initial EPE error is 36.72, and the initialization module reduces the initial error to 10.96, reducing the subsequent optimizer's training difficulty and achieving higher accuracy with fewer iterations.}
	\label{fig_init}
\end{figure}

\begin{table}[htbp]
	\centering
	\caption{\textbf{Runtime Analysis of Different Optimizers.} Perform 100 consecutive inferences on a $375\times1242$ image and take the average. The GPU used is RTX4090.}
	\setlength{\tabcolsep}{3.6pt}
	\begin{tabular}{lccc}
		\toprule
		Optimizer & \multicolumn{1}{l}{Single Iteration Time} & \multicolumn{1}{l}{Refine Iterations} & \multicolumn{1}{l}{Total Time} \\
		\midrule
		GIR (ours)   & 3.26ms & 6     & 19.56ms \\
		RNN   & 1.92ms & 12    & 23.04ms \\
		GRU   & 1.62ms & 12    & 19.44ms \\
		\bottomrule
	\end{tabular}%
	\label{tab:time}%
\end{table}%

\textbf{Iterative Optimizer:} 
We compared the performance between the most commonly used GRU optimizer, RNN optimizer in SEA-RAFT\cite{wang2024sea}, and the GIR optimizer proposed in this paper. As shown in rows E, F, and G in Table~\ref{tab:ablation}, all optimizers have undergone 6 iterations, and GIR has significant advantages in Sintel scenarios and $Mid$ metrics. Moreover, we also reported the running time of each optimizer in the complete method in Table~\ref{tab:time}. The optimizer proposed in this paper did not significantly increase computational overhead under standard iteration numbers.

\textbf{Initialize Displacement Field:} 
As shown in row H of Table~\ref{tab:ablation}, the initialization module significantly improved the performance of optical flow and slightly enhanced $Mid$ performance, proving the effectiveness of the initialization module.

We also demonstrated a more detailed working process of the initialization module. As shown in Fig.~\ref{fig_init}, the initialization module reduced the initial optical flow EPE from 36.72 to 10.96, alleviating the task alienation of the optimizer at different iteration stages, reducing the overall number of iterations, and improving accuracy.

\begin{table}[htbp]
	\centering
	\caption{\textbf{Motion-in-depth estimation on K-40.} The best among all are bolded, and the second best are underlined. Our method outperforms the monocular baselines by a large margin.}
	\setlength{\tabcolsep}{3.9pt}
	\begin{tabular}{lcccc}
		\toprule
		Method & Input  & Training set & \multicolumn{1}{l}{$Mid \downarrow$} & \multicolumn{1}{l}{Time/s} \\
		\midrule
		OSF\cite{menze_object_2015}   & Stereo & K-160 & 115   & 3000 \\
		PRSM\cite{vogel_3d_2015}  & Stereo & K-160 & 124   & 300 \\
		Hur \& Roth\cite{hur2020self}   & Mono & K-160 & 115.13   & 0.1 \\
		Binary TTC\cite{badki_binary_2021} & Mono  & K-160 & 73.55 & 2.2 \\
		Optical expansion\cite{yang_upgrading_2020} & Mono  & K-160 & 75    & 0.2 \\
		Scale-flow \cite{ling2022scale}  & Mono  & K-160 & 48.9 & 0.2 \\
		TPCV \cite{ling2023learning}  & Mono  & K-160 & \underline{42.84} & 0.2 \\
		ScaleFlow++(ours)  & Mono  & K-160 & \textbf{38.44} & 0.2 \\
		
		\bottomrule
	\end{tabular}%
	\label{tab:t1}%
\end{table}%

\begin{figure*}[!t]
	\centering
	\includegraphics[width=6.4in]{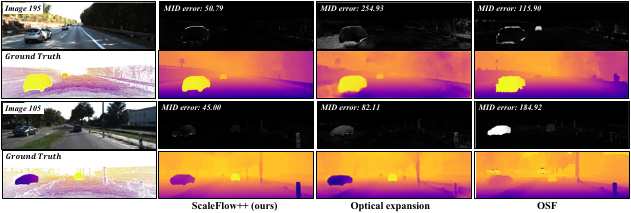}
	\caption{Motion-in-depth results for images ``105'' and ``195'' in the K-40 set. For each of them, Up: input images and the error map, where the whiter the pixels, the greater the error. Down: ground truth and visualization of motion-in-depth. Our method is much more accurate than other methods.}
	\label{fig_MID}
\end{figure*}

\begin{figure}[!t]
	\centering
	\includegraphics[width=3.3in]{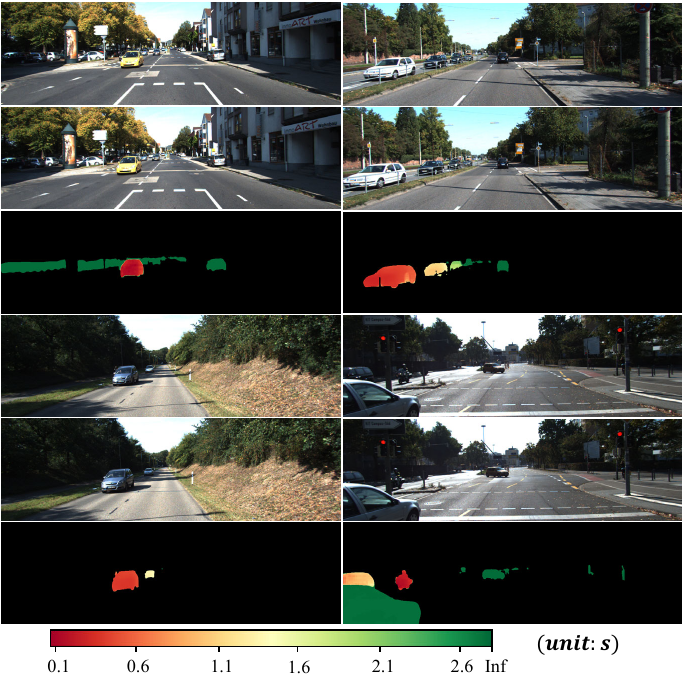}
	\caption{Visualization of foreground target collision time.
		We visualized some of the TTC results in K-40; from top to bottom are two consecutive frames and the visualized collision time.
		The foreground mask is provided by the real-time segmentation algorithm Mseg\cite{MSeg_2020_CVPR}.}
	\label{fig_ttc}
\end{figure}

\subsection{Motion-in-depth (MID)}\label{ex:mid}
MID is the core indicator of monocular 3D flow methods. In order to prove the superiority of ScaleRAFT in estimating MID tasks, we compared ScaleFlow++ with the stereo-based traditional scene flow methods, state-of-the-art MID methods and scene flow methods on K-40. We use the same criteria as predecessors\cite{yang_upgrading_2020} to define the loss:
\begin{equation}
	Mid = ||log(f_3)-log(\bm{\tau}_{gt})||_1\cdot10^4
	\label{eq:eqmid}
\end{equation}
where $\bm{\tau}_{gt}$ is the ground truth of MID.

We first compare with the stereo-based traditional approach: OSF and PRSM decompose the pixels into rigid blocks, and iteratively update the optical flow and depth field based on rigidity assumptions and hand-designed regularization terms.To get the MID result, we divide the depth of the second frame by the depth of the first frame.
The experimental results show that our error is much smaller (115 v.s. 38.44) and in less time.

Then we compare with the state-of-the-art MID methods Optical expansion, TPCV and Binary TTC. As shown in Table~\ref{tab:t1}, ScaleFlow++ still has apparent advantages (42.84 v.s. 38.44). 

\textbf{Application example:} We present an application example of MID in downstream 3D tasks: time-to-collision (TTC).

Time-to-collision estimation is a crucial technique in path planning for autonomous robots\cite{lee1976theory,byrne_expansion_2009,marinho_guaranteed_2018,mori_first_2013}, which describes the collision time between moving objects and the observer plane. TTC can be estimated indirectly by MID $\bm{\tau}$:
\begin{equation}
	TTC = \frac{Z}{Z-Z'}T = \frac{T}{1-\frac{Z'}{Z}}=\frac{T}{1-\bm{\tau}}
	\label{eq:eq17}
\end{equation}
where $T$ is the sampling interval between two consecutive frames, for KITTI $T=0.1s$, $Z$ and $Z'$ are the object's depth in frame1 and frame2.

We demonstrate the TTC performance of ScaleFlow++ in Fig.~\ref{fig_ttc}. Our method can easily distinguish the collision sequence and approximate collision time of different moving objects.

\subsection{Optical Flow} \label{ex:op}
Results are shown in Tab.~\ref{tab:opticalflow}. Compared to the previous version of Scale-flow, ScaleFlow++ has improved the overall performance by at least 18\% and 19\% on Sintel and KITTI, respectively, demonstrating the effectiveness of the improvements proposed in this paper.

Moreover, compared with existing state-of-the-art optical flow methods, ScaleFlow++ is also very competitive. On the Sintel test, it leads most RAFT-based methods and achieves even better performance on the KITTI test.

\begin{table*}
	\centering
	\caption{\textbf{State-of-the-art published methods on KITTI scene flow benchmark.} RGB-D means depth and monocular information, and Mono means monocular information. D1, D2, Fl, and SF is the percentage of disparity, optical flow and scene flow outliers. -bg,-fg and -all represent the percentage of outliers averaged only over background regions, foreground regions and overall ground truth pixels.
		The best among the same group are bolded, and the second best are underlined.	
		Our monocular method performs quite well, not only far ahead of similar monocular methods, but also outperforming RGBD-based methods in foreground scene flow estimation, which proves that the cross-scale matching can accurately estimate the 3D motion.}
	\begin{tabular}{llcccccccccccc}
		\toprule
		Method & Typev & \multicolumn{1}{l}{D1-bg} & \multicolumn{1}{l}{D1-fg} & \multicolumn{1}{l}{D1-all} & \multicolumn{1}{l}{D2-bg} & \multicolumn{1}{l}{D2-fg} & \multicolumn{1}{l}{D2-all} & \multicolumn{1}{l}{Fl-bg} & \multicolumn{1}{l}{Fl-fg} & \multicolumn{1}{l}{Fl-all} & \multicolumn{1}{l}{SF-bg} & \multicolumn{1}{l}{SF-fg} & \multicolumn{1}{l}{SF-all} \\
		\midrule
		
		CamLiFlow\cite{liu2022camliflow} & \multirow{4}[2]{*}{RGBD} & 1.48  & 3.46  & 1.81  & - & -  & 3.19 & - & -  & 4.05 & - & - & 5.62 \\
		CamLiFlow-RBO\cite{liu2022camliflow} &   & 1.48  & 3.46  & 1.81  & 1.92 & 8.14  & 2.95 & 2.31 & 7.04  & 3.1 & 2.87 & 12.23 & 4.43 \\
		RigidMask+ISF\cite{yang_learning_2021} &   & 1.53  & 3.65  & 1.89  & 2.09  & 8.92  & 3.23  & 2.63  & 7.85  & 3.5   & 3.25  & 13.08 & 4.89 \\
		RAFT3D\cite{teed_raft-3d_2021} &   & 1.48  & 3.46  & 1.81  & 2.51  & 9.46  & 3.67  & 3.39  & 8.79  & 4.29  & 4.27  & 13.27 & 5.77 \\
		\midrule
		Optical expansion\cite{yang_upgrading_2020} & \multirow{5}[2]{*}{Mono}  & 1.48  & 3.46  & 1.81  & 3.39  & 8.54  & 4.25  & 5.83  & 8.66  & 6.3   & 7.06  & 13.44 & 8.12 \\
		Binary TTC\cite{badki_binary_2021} &    & 1.48  & 3.46  & 1.81  & 3.84  & 9.39  & 4.76  & 5.84  & 8.67  & 6.31  & 7.45  & 13.74 & 8.5 \\
		TPCV\cite{ling2023learning} &   & 1.48  & 3.46  & 1.81  & \underline{2.29}  & \textbf{7.63}  & \underline{3.18}  & \underline{4.53}  &\textbf{5.52} & \underline{4.69}  & \underline{5.34} & \textbf{10.60} & \underline{6.21} \\
		Scale-flow\cite{ling2022scale}  &    & 1.48  & 3.46  & 1.81  & 2.55  & 8.24 & 3.50  & 5.24  & 5.71 & 5.32  & 6.06  & 11.32 & 6.94\\
		ScaleFlow++ (ours) &    & 1.48  & 3.46  & 1.81  & \textbf{2.13}  & \underline{8.02} & \textbf{3.11}  & \textbf{3.94}& \underline{5.59}& \textbf{4.21}  & \textbf{4.81}  & \underline{10.69} & \textbf{5.79} \\
		
		\bottomrule
	\end{tabular}%
	\label{tab:t5}%
\end{table*}

\begin{table}[htbp]
	\centering
	\caption{\textbf{Compare optical flow performance with other state-of-the-art methods.} The best among all are bolded, and the second best are underlined. * denotes the use of TartanAir\cite{wang2020tartanair} as additional training data.}
	\setlength{\tabcolsep}{2.4pt}
	\begin{tabular}{llccccc}
		\toprule
		&       & \multicolumn{2}{c}{Sintel-test} &       & \multicolumn{1}{l}{K15-test} &  \\
		\midrule
		Base  & Method & Clean$\downarrow$ & Final$\downarrow$ & Fl-all$\downarrow$ & Fl-bg$\downarrow$ & Fl-fg$\downarrow$\\
		\midrule
		\multirow{7}[2]{*}{RAFT} & SEA-RAFT(M)* & 1.44  & 2.86  & 4.64  & 4.47  & \underline{5.49}  \\
		& SEA-RAFT(L)* & 1.31 & 2.60  & \underline{4.30}  & \underline{4.08}  & \textbf{5.37} \\
		& RAFT\cite{teed_raft_2020}  & 1.61  & 2.86  & 5.10  & 4.74  & 6.87  \\
		& CRAFT\cite{sui2022craft} & 1.45  & 2.42 & 4.79  & 4.58  & 5.85  \\
		& MS-RAFT\cite{jahedi2022high} & 1.37  & 2.67 & 4.58  & 4.88  & 6.38  \\
		& GMA\cite{jiang2021learning}   & 1.39  & 2.47  & 5.15  &    -   &  -\\
		& Scale-flow\cite{ling2022scale} & 1.88  & 3.16  & 5.32  & 5.24  & 5.71  \\
		& ScaleFlow++(ours) & 1.33  & 2.77  & \textbf{4.21} & \textbf{3.94} & 5.59 \\
		\midrule
		\multirow{4}[2]{*}{Transformer} & GMFlowNet\cite{zhao2022global}& 1.39  & 2.65  & 4.79  & 4.39  & 6.84  \\
		& GMFlow\cite{xu2022gmflow} & 1.74  & 2.90  & 9.32  & 9.67  & 7.57  \\
		& FlowFormer\cite{huang2022flowformer} & \underline{1.20}  & \textbf{2.12} & 4.68  & 4.37  & 6.18  \\
		& GMFlow++\cite{xu2023unifying} & \textbf{1.03} & \underline{2.37}  & 4.49 & 4.27 & 5.60\\
		\bottomrule
	\end{tabular}%
	\label{tab:opticalflow}%
\end{table}%

\begin{figure}[!t]
	\centering
	\includegraphics[width=3.36in]{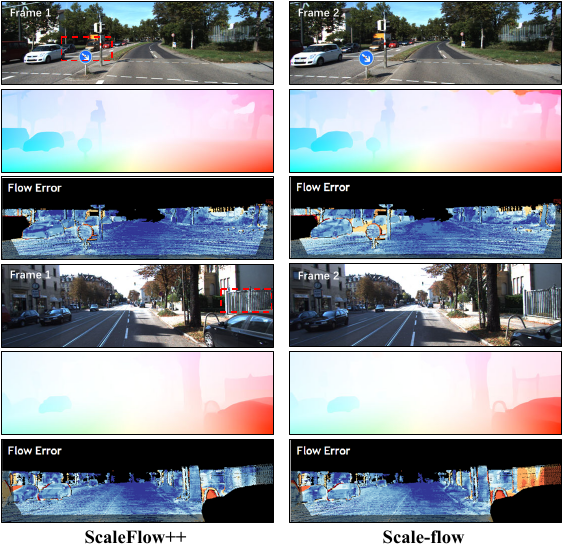}
	\caption{\textbf{Visualization results on the KITTI test.} In each group, from top to bottom, there are two consecutive frames of images, the visualized optical flow, and the visualized optical flow error map. The redder the color of the error map, the greater the error. Observing the red box area in the image, our improved method better optimizes the background over a large area. }
	\label{fig_bg}
\end{figure}

\subsection{Scene Flow} \label{ex:scene}
Correctly estimating the 3D motion of an object is crucial for the robot path planning and action prediction task in a dynamic environment\cite{kong2022human,menze_object_2015}.
Scene flow consists of optical flow and depth changes of matched pixels between two frames.
In order to unify the evaluation standards with the most advanced methods, we use GA-Net\cite{zhang2019ga} to obtain the depth of the first frame $D1$ and calculate the depth change through MID $\bm{\tau}$, where $D2 = \bm{\tau} * D1$. The model was trained on the  K-200 and tested on the KITTI public scene flow evaluation website.

We first compare ScaleFlow++ with monocular methods, as shown in Table~\ref{tab:t5}, ScaleFlow++ leads in all monocular methods with significant advantages, consistent with the previous superior performance in motion-in-depth tasks. It proves the effectiveness of our ScaleFlow++ framework in the field of monocular 3D motion estimation. 

We also compare ScaleFlow++ with the state-of-the-art RGBD-based methods RAFT3D, CamLiFlow, and RigidMask+ISF. 
As shown in Tab. \ref{tab:t5}, RGBD-based methods tend to get smaller Fl-bg errors, which is thanks to the aid of depth-based rigid assumption to optimize the background optical flow. Nonetheless, our monocular method still achieved competitive results in the core indicator SF-all and significant advantages in foreground estimation (SF-fg) compared to the RGBD-based method (10.69 v.s. 12.23).

\begin{figure*}[!t]
	\centering
	\includegraphics[width=6.4in]{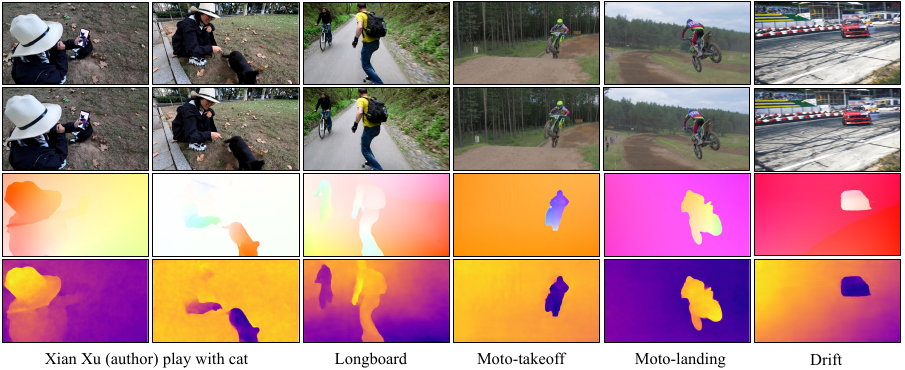}
	\caption{Self-supervised generalization. ScaleFlow++ has demonstrated stunning results in natural scenes that have never been seen before, demonstrating that our method is robust and easy to generalize.}
	\label{fig_sgrw}
\end{figure*}

\subsection{Motion-in-depth on KITTI Test}
Since the KITTI testing platform does not directly indicate motion-in-depth, we indirectly measure the accuracy of motion-in-depth based on the depth error growth (DEG) rate.

\begin{equation}
	\textrm{DEG}=\frac{\textrm{D2}-\textrm{D1}}{\textrm{D1}}
	\label{eq:DEG}
\end{equation}
where D1 and D2 are the depth outlier rates in the first and second frames, respectively. DEG describes how much error is added to the depth corresponding to the second frame based on the depth estimation of the first frame.
\begin{table}[htbp]
	\centering
	\caption{\textbf{Comparison of Growth Rates of Deep Motion Error on KITTI Test.}}
	\begin{tabular}{lcccc}
		\toprule
		Method & \multicolumn{1}{l}{Input} & \multicolumn{1}{l}{DEG-bg$\downarrow$} & \multicolumn{1}{l}{DEG-fg$\downarrow$} & \multicolumn{1}{l}{DEG-all$\downarrow$} \\
		\midrule
		CamLiFlow-RBO & \multirow{3}[2]{*}{RGBD} & 0.30  & 1.35  & 0.63  \\
		RigidMask+ISF &       & 0.37  & 1.59  & 0.71  \\
		RAFT3D &       & 0.70  & 1.73  & 1.03  \\
		\midrule
		Optical expansion & \multirow{5}[2]{*}{Mono} & 1.29  & 1.47  & 1.35  \\
		Binary TTC &       & 1.59  & 1.71  & 1.63  \\
		TPCV  &       & \underline{0.55}  & \textbf{1.21} & \underline{0.76}  \\
		Scale-flow &       & 0.72  & 1.38  & 0.93  \\
		ScaleFlow++ &       & \textbf{0.44} & \underline{1.32}  & \textbf{0.72} \\
		\bottomrule
	\end{tabular}%
	\label{tab:midtest}%
\end{table}%

As shown in Tab.~\ref{tab:midtest}, our method outperforms all monocular methods in DEG-all and is highly competitive even compared to RGBD methods. In addition, the performance of ScaleFlow++ is particularly impressive in the evaluation of the background area, mainly due to the better global view provided by the GIR module to optimize the background.

\subsection{Generalization in Real World} \label{sec66}
To assist readers in better applying ScaleFlow++, we trained generalization weights on 60000 real-world image pairs using a 3D flow self-supervised generalization method, ADFactory\cite{ling2024adfactory}.

As shown in Fig.~\ref{fig_sgrw}, whether it is optical flow or MID, the ScaleFlow++ trained by self-supervision has shown excellent generalization results in unseen scenarios. We hope that our work can help methods in fields such as human pose recognition, action prediction, and autonomous driving better perceive the three-dimensional world.

\section{Conclusion}
In this article, we propose a new cross-scale recurrent all-pairs field transform for 3D motion estimation (ScaleFlow++), which extends the optical flow matching from the original image-to-image to image-to-scale-space, reducing the interference of scale changes on optical flow matching. Moreover, matching in scale space can also obtain depth motion clues, helping 3D flow methods break away from depth dependence and regression dependence.  We also propose a global iterative refinement (GIR) module and an initialization module to address the inherent problems of small network perception range and task alienation in iterative methods. These modules enable the network to perceive global motion and achieve more robust optimizer learning in fewer iterations. The comprehensive evaluation of different datasets shows that ScaleFlow++ performs excellently in various 3D tasks, especially in scenes with large scale changes (driving scenes). We hope this exciting result draws people to pay more attention to 3D motion estimation based on a monocular camera.

\bibliographystyle{IEEEtran}
\bibliography{Mybib}

\end{document}